\documentclass[twocolumn,conference, letterpaper]{ieeeconf}
\usepackage[T1]{fontenc}
\usepackage[latin9]{inputenc}
\usepackage{color}
\usepackage{array}
\usepackage{booktabs}
\usepackage{multirow}
\usepackage{amstext}
\usepackage[unicode=true,
 bookmarks=true,bookmarksnumbered=true,bookmarksopen=true,bookmarksopenlevel=1,
 breaklinks=false,pdfborder={0 0 0},pdfborderstyle={},backref=false,colorlinks=false]
 {hyperref}
\hypersetup{pdftitle={Your Title},
 pdfauthor={Your Name},
 pdfpagelayout=OneColumn, pdfnewwindow=true, pdfstartview=XYZ, plainpages=false}

\makeatletter

\providecommand{\tabularnewline}{\\}

\usepackage[caption=false,font=footnotesize]{subfig}

\usepackage{graphicx}
\usepackage{import}

\usepackage{cite}

\usepackage{balance}

\usepackage{resizegather}

\IEEEoverridecommandlockouts 

\makeatother

\begin{document}
\title{Single-Shot Pose Estimation of Surgical Robot Instruments' Shafts
from Monocular Endoscopic Images}
\author{Masakazu~Yoshimura\emph{,} Murilo~M.~Marinho, Kanako~Harada, Mamoru~Mitsuishi\thanks{This
work was supported by JSPS KAKENHI Grant Number 19K14935. \emph{(Corresponding
author:} Murilo M. Marinho)}\thanks{Masakazu Yoshimura, Murilo M.
Marinho, Kanako Harada, and Mamoru Mitsuishi are with the Department
of Mechanical Engineering, the University of Tokyo, Tokyo, Japan.
\texttt{Emails:\{m.yoshimura, murilo, kanako, mamoru\}@nml.t.u-tokyo.ac.jp}.
}}
\maketitle
\begin{abstract}
Surgical robots are used to perform minimally invasive surgery and
alleviate much of the burden imposed on surgeons. Our group has developed
a surgical robot to aid in the removal of tumors at the base of the
skull via access through the nostrils. To avoid injuring the patients,
a collision-avoidance algorithm that depends on having an accurate
model for the poses of the instruments' shafts is used. Given that
the model's parameters can change over time owing to interactions
between instruments and other disturbances, the online estimation
of the poses of the instrument's shaft is essential. In this work,
we propose a new method to estimate the pose of the surgical instruments'
shafts using a monocular endoscope. Our method is based on the use
of an automatically annotated training dataset and an improved pose-estimation
deep-learning architecture. In preliminary experiments, we show that
our method can surpass state of the art vision-based marker-less pose
estimation techniques (providing an error decrease of 55\% in position
estimation, 64\% in pitch, and 69\% in yaw) by using artificial images.
\end{abstract}

\section{Introduction}

Robot-assisted surgery has many advantages over manual surgery. In
robot-assisted surgery, surgeons\textquoteright{} hand tremors are
filtered, and they can move robotic instruments more precisely. Moreover,
surgeons can efficiently perform dexterous manipulation of robotic
instruments that have multiple degrees of freedom at the tip, whereas
hand-held surgical instruments have limited dexterity. Considering
these advantages, we are developing a versatile robot, called SmartArm
\cite{Marinho2020}, designed with a focus on procedures in deep and
narrow spaces.

\begin{figure}[t]
\centering

\def\svgwidth{1.0\columnwidth}

\scriptsize
\begingroup%
  \makeatletter%
  \providecommand\color[2][]{%
    \errmessage{(Inkscape) Color is used for the text in Inkscape, but the package 'color.sty' is not loaded}%
    \renewcommand\color[2][]{}%
  }%
  \providecommand\transparent[1]{%
    \errmessage{(Inkscape) Transparency is used (non-zero) for the text in Inkscape, but the package 'transparent.sty' is not loaded}%
    \renewcommand\transparent[1]{}%
  }%
  \providecommand\rotatebox[2]{#2}%
  \newcommand*\fsize{\dimexpr\f@size pt\relax}%
  \newcommand*\lineheight[1]{\fontsize{\fsize}{#1\fsize}\selectfont}%
  \ifx\svgwidth\undefined%
    \setlength{\unitlength}{627.51690674bp}%
    \ifx\svgscale\undefined%
      \relax%
    \else%
      \setlength{\unitlength}{\unitlength * \real{\svgscale}}%
    \fi%
  \else%
    \setlength{\unitlength}{\svgwidth}%
  \fi%
  \global\let\svgwidth\undefined%
  \global\let\svgscale\undefined%
  \makeatother%
  \begin{picture}(1,0.47697602)%
    \lineheight{1}%
    \setlength\tabcolsep{0pt}%
    \put(0,0){\includegraphics[width=\unitlength,page=1]{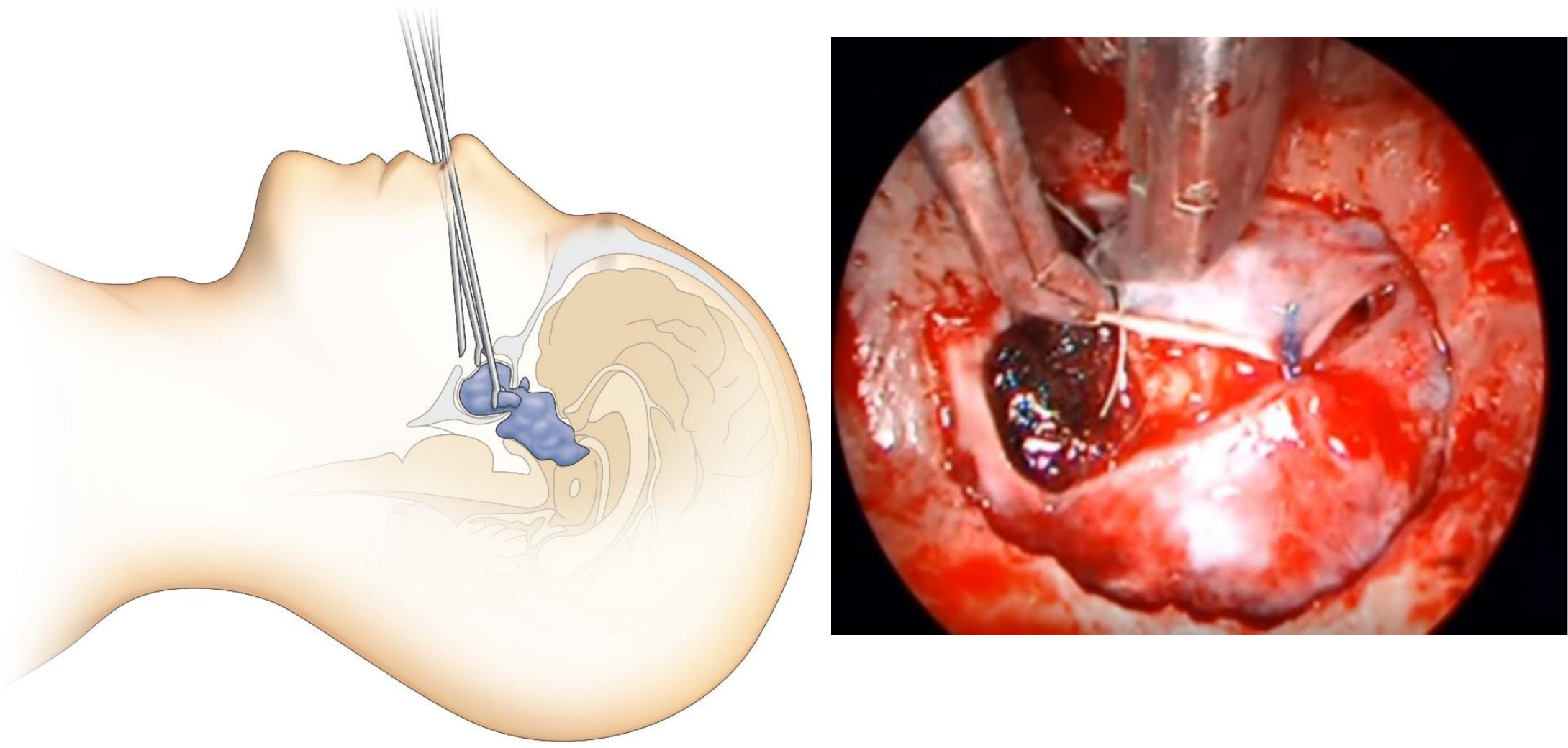}}%
    \put(0.78874907,0.0290235){\color[rgb]{0,0,0}\makebox(0,0)[lt]{\lineheight{1.25}\smash{\begin{tabular}[t]{l}Dura mater\end{tabular}}}}%
    \put(0.0501341,0.17456188){\color[rgb]{0,0,0}\makebox(0,0)[lt]{\lineheight{1.25}\smash{\begin{tabular}[t]{l}Pituitary gland\end{tabular}}}}%
    \put(0,0){\includegraphics[width=\unitlength,page=2]{transsphenoidal_surgery_v2.pdf}}%
  \end{picture}%
\endgroup%

\caption{\label{fig:transsphenoidal_surgery} Transsphenoidal surgery (left)
and the endoscopic image during dura mater suturing (right).}
\end{figure}

One of the applications of the SmartArm is endonasal surgery, in which
transsphenoidal surgery is one goal. Transsphenoidal surgery is a
procedure to remove tumors of the pituitary gland or other tumors
at the skull base, as shown in Fig.~\ref{fig:transsphenoidal_surgery}.
In robot-assisted surgeries in constrained workspaces such as the
nasal cavity, surgeons have restricted vision to the region near the
surgical instruments' tips. To autonomously prevent collisions between
the robots and surrounding tissues, we have developed a virtual-fixtures
framework based on the kinematic model of the robots \cite{marinho2019dynamic}.

However, even with the careful offline calibration of the robots'
parameters and in a considerably controlled environment, we can still
observe a mismatch of a few millimeters between the kinematic model
and the absolute pose\footnote{Pose means combined position and orientation.}
of the tools. Moreover, even if a perfect offline calibration of the
robots were possible, disturbances such as changes in temperature
and the interactions of the tools with tissues cause the robots' parameters
to change over time. This mismatch between the calculated pose and
the absolute pose of the robots' instruments are even more pronounced
in cable-driven robots, as reported in related literature \cite{reiter2014appearance,Moccia2020}
using the da Vinci Surgical System (Intuitive Surgical, USA), which
is a robotic system routinely used for surgeries \emph{in-vivo}.

The reason why this mismatch is not apparent for users in teleoperation
\cite{Marinho2019} is that the human operator ``closes the loop''
by using their vision. As we move towards (semi-)automation of surgical
tasks, a system to ``close the loop'' and provide online calibration
of the tool parameters is paramount. Moreover, it is important to
achieve a highly-accurate parameter calibration given that the effectiveness
of the collision avoidance depends heavily on the accuracy of the
robots' model.

\subsection{Related works}

Many strategies have been proposed to estimate the pose of surgical
instruments. Some require added sensors, such as ultrasound \cite{ren2012tubular}
or electromagnetic trackers \cite{kim2015real}. Such a requirement
makes it difficult to impact existing operating rooms, and the sensors
themselves have limitations that should be considered. For instance,
the propagation of ultrasound is affected by its medium, and electromagnetic
sensors are affected by surrounding metallic materials and magnetic
fields.

An alternative is to use a stereo camera. Allan \emph{et al.} \cite{allan2015image}
calculated the pose of surgical instruments from the parallax of a
stereo laparoscope using a particle filter and optical flow after
semantic segmentation. Other approaches, based on correlating the
2D images to 3D templates. Baek \emph{et al.} \cite{Baek2014} used
particle filtering and kinematic data to track instruments of a microsurgical
robotic system. Moccia \emph{et al}. \cite{Moccia2020} used feature
matching and an extended Kalman filter to estimate and track the pose
of the da Vinci's instruments. However, endonasal procedures require
thin (3 mm) endoscopes, that are not currently available on the market.

Printable markers have been proposed by Gadwe \emph{et al.} \cite{gadwe2018real},
but that would also require modifying the surgical instruments.

Based on the above, we aim to estimate the instruments' pose with
a monocular endoscope. A myriad of works have explored the estimation
of instruments' pose from monocular images. For example, Reiter \emph{et
al.} \cite{reiter2014appearance} estimated the da Vinci's instruments'
pose using keypoint information. Ye \emph{et al}. \cite{ye2016real}
studied keypoint extraction and parts-based template matching to estimate
tool pose even in occluded situations. Zhou and Payandeh \cite{Zhou2014}
estimated the pose from the position of straight contour lines and
the tip on the image. Other groups \cite{hao2018vision,allan20183}
have proposed the estimation by two-step approaches, using the segmentation
and template matching method plus object tracking. In this category
of instrument pose estimation, errors larger than 2.8 mm in position
and 4.8$\ensuremath{^{\circ}}$ in rotation are reported. Accurate
pose estimation of surgical instruments from monocular images is still
a challenging task.

Deep learning technology is being rapidly developed and pose estimation
of objects from monocular images with end-to-end learning and inference
has been studied in the past few years. For example, Sundermeyer \emph{et
al}. \cite{sundermeyer2018implicit} estimated the pose of real objects
from monocular images using a training dataset composed of artificial
images rendered from 3D simulated objects. In their approach, the
objects are first detected using single-shot detection (SSD) \cite{liu2016ssd};
then, the pose is estimated using an auto-encoder trained with artificial
images. In another work, \emph{Kehl et al}. \cite{kehl2017ssd} proposed
the SSD-6D method, in which the object pose was estimated directly
as the output of the SSD together with the bounding boxes of objects
on the images. To the best of our knowledge, these methods have not
been applied to the pose estimation of surgical tools.

\subsection{Limitations of current methods}

In a pilot evaluation, we attempted to use the SSD-6D methodology
directly to estimate the pose of the shaft of our surgical instruments
in our robotic setup. We identified two limitations of the current
methods that we address in this work:
\begin{enumerate}
\item In SSD-6D, the pose estimation is regarded as a classification problem
by discretizing the pose-space. For example, rotations were classified
in five-degree steps, that are not precise enough for the pose estimation
of surgical instruments.
\item It is assumed that the objects are reasonably far from the camera,
so the images are mostly unaffected by perspective distortions. This
assumption does not hold true in the case of endonasal endoscopic
images because the instruments are close to the lens, and endoscopes
usually have wide perspective angles. This causes the object's shape
to be deformed in the endoscopic image owing to perspective distortions
(closer parts of the object look larger than farther parts of the
same object).
\end{enumerate}

\subsection{Statement of contributions}

With prior literature in mind, our goal in this work is to estimate
the pose of surgical instruments from monocular endoscopic images
using deep learning end-to-end. Our use-case breaks some of the assumptions
of earlier methods because the appearance of the instrument' changes
according to the distance from the endoscope due to perspective distortions.
Moreover, only the tips of the instruments can be seen on endoscopic
images. To overcome these issues, we propose a new deep learning architecture.
It is an improved architecture of the SSD-6D \cite{kehl2017ssd} that
performs regression instead of classification. The network also responds
well to occlusion and a varying number of instruments. The generation
of training data also relies partially on artificial data augmentation,
using computer graphics (CG) rendering software. Our results show
that we can accurately estimate the pose of surgical instruments from
real endoscopic images. To the best of our knowledge, the accuracy
of our method is the state of the art in vision-based and marker-less
pose estimation of surgical instruments.

\section{Problem statement}

\begin{figure}[h]
\centering

\def\svgwidth{1.0\columnwidth}

\scriptsize
\begingroup%
  \makeatletter%
  \providecommand\color[2][]{%
    \errmessage{(Inkscape) Color is used for the text in Inkscape, but the package 'color.sty' is not loaded}%
    \renewcommand\color[2][]{}%
  }%
  \providecommand\transparent[1]{%
    \errmessage{(Inkscape) Transparency is used (non-zero) for the text in Inkscape, but the package 'transparent.sty' is not loaded}%
    \renewcommand\transparent[1]{}%
  }%
  \providecommand\rotatebox[2]{#2}%
  \newcommand*\fsize{\dimexpr\f@size pt\relax}%
  \newcommand*\lineheight[1]{\fontsize{\fsize}{#1\fsize}\selectfont}%
  \ifx\svgwidth\undefined%
    \setlength{\unitlength}{648.7199707bp}%
    \ifx\svgscale\undefined%
      \relax%
    \else%
      \setlength{\unitlength}{\unitlength * \real{\svgscale}}%
    \fi%
  \else%
    \setlength{\unitlength}{\svgwidth}%
  \fi%
  \global\let\svgwidth\undefined%
  \global\let\svgscale\undefined%
  \makeatother%
  \begin{picture}(1,0.48316685)%
    \lineheight{1}%
    \setlength\tabcolsep{0pt}%
    \put(0,0){\includegraphics[width=\unitlength,page=1]{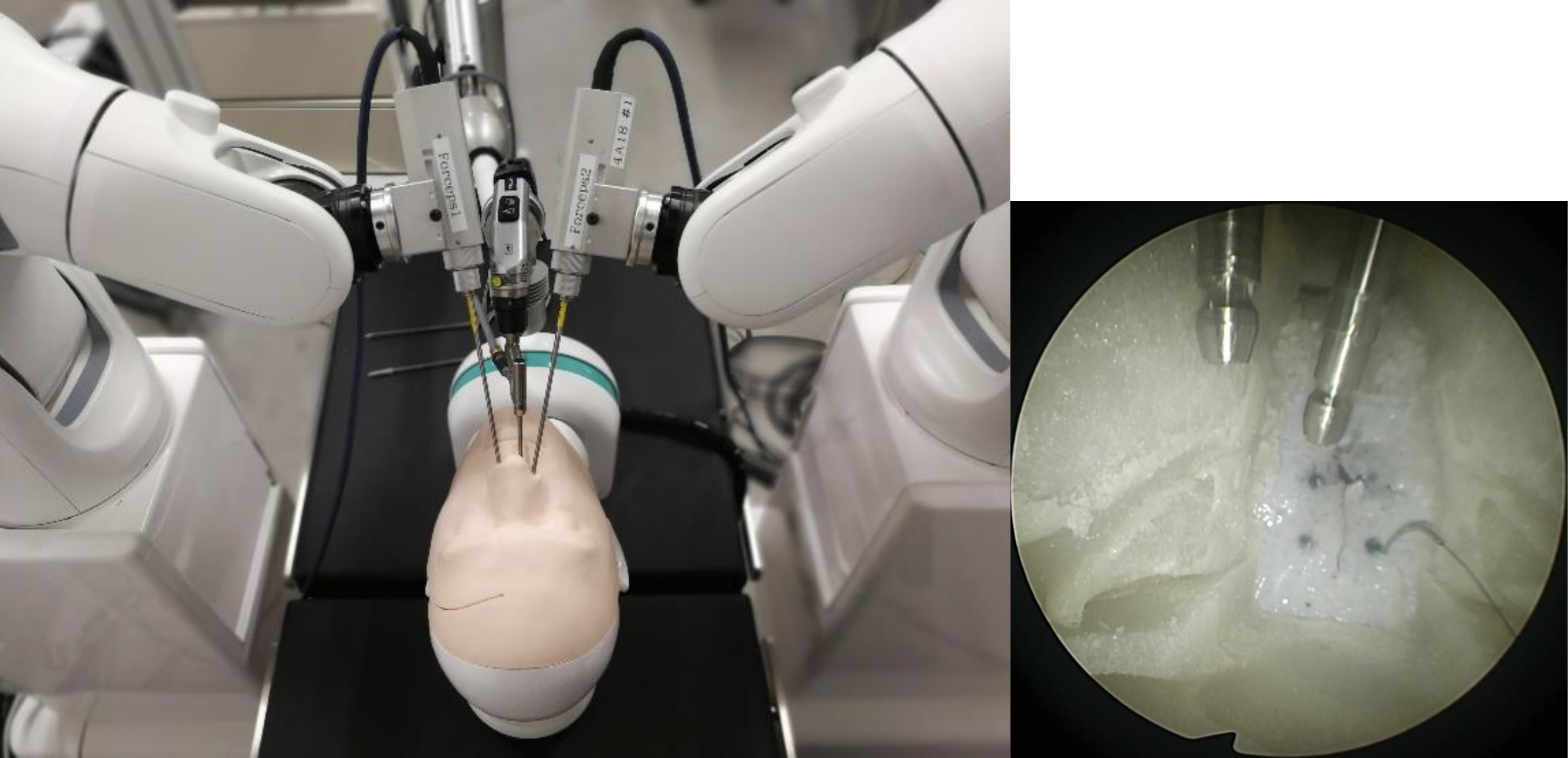}}%
    \put(0.73005304,0.40251576){\color[rgb]{0,0,0}\makebox(0,0)[lt]{\lineheight{1.25}\smash{\begin{tabular}[t]{l}Instruments\end{tabular}}}}%
    \put(0.72715502,0.44305713){\color[rgb]{0,0,0}\makebox(0,0)[lt]{\lineheight{1.25}\smash{\begin{tabular}[t]{l}Monocular endoscope\end{tabular}}}}%
    \put(0,0){\includegraphics[width=\unitlength,page=2]{robot.pdf}}%
    \put(0.70622575,0.33042239){\color[rgb]{0,0,0}\makebox(0,0)[lt]{\lineheight{1.25}\smash{\begin{tabular}[t]{l}Simplified instruments \end{tabular}}}}%
    \put(0,0){\includegraphics[width=\unitlength,page=3]{robot.pdf}}%
    \put(0.65951248,0.05253798){\color[rgb]{0,0,0}\makebox(0,0)[lt]{\lineheight{1.25}\smash{\begin{tabular}[t]{l}Artificial membrane\end{tabular}}}}%
    \put(0,0){\includegraphics[width=\unitlength,page=4]{robot.pdf}}%
  \end{picture}%
\endgroup%

\caption{\label{fig:statement} Robot setup (left) and its inner view from
a 70$^{\circ}$ endoscope (right).}
\end{figure}

Given the robot-aided endonasal surgery setup, as shown in Fig.~\ref{fig:statement},
let there be two robotic arms (SmartArm \cite{Marinho2020}) with
instruments as their end effectors. Those instruments are inserted
through the nostrils of an anatomically-realistic head model (BionicBrain
\cite{Masuda2019}). Images are obtained through a high-definition
endoscopic system (Endoarm, Olympus, Japan) with a 70$^{\circ}$ endoscope
with perspective angle of 95$^{\circ}$.

For this work, suppose that we are only interested in finding the
pose of the instrument's shaft because this information is enough
for appropriate collision avoidance in the context of endonasal surgery
\cite{marinho2019dynamic}. The reference frames of the instruments
are attached to their distal tip, and the $z$-axis is along the instruments'
shaft. In this context, our target is to estimate the pose of the
intruments' shafts relative to the monocular endoscope. In future
work, our vision will be feedback this information to the robot controller,
to adapt the robot's parameters, and to increase the safety and precision
of the robot-aided procedure.

\section{Methodology}

This paper proposes a new method of pose estimation of the surgical
instrument's shafts from monocular endoscopic images. In this section,
we first describe how to create a training dataset composed of artificial
CG images. Second, we explain our proposed network architecture for
pose estimation. Lastly, we explain the data augmentation approach
and the loss function.

\subsection{Training dataset}

\begin{figure*}[tbh]
\centering

\def\svgwidth{2.0\columnwidth}

\scriptsize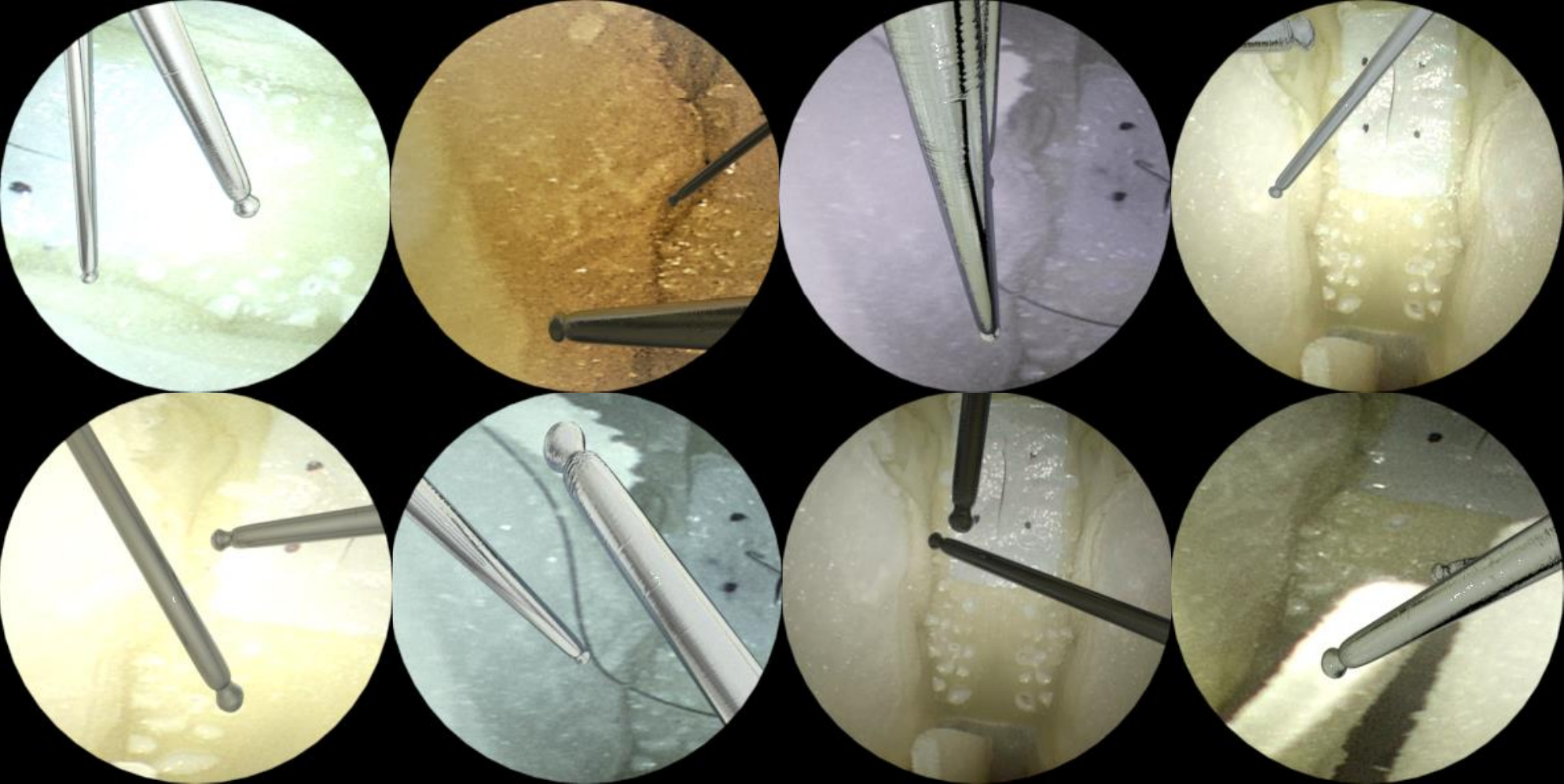

\caption{\label{fig:sample_artificial_images} Samples of artificial images.}
\end{figure*}

Deep convolutional neural networks (DCNNs) are known to need a large
amount of training data to avoid over-fitting, and manual annotation
is a time-consuming and error-prone task. In multi-task learning,
even more complicated annotation is required. For instance, in SSD-6D
and this work, the semantic annotation, the bounding boxes, and the
precise pose of the shafts are necessary. To address this issue, we
partially relied on the automatic annotation of CG images using an
open-source rendering software (Blender, Blender Foundation, Netherlands),
using a rendering pipeline based on \cite{Marinho2017}.

To generate the CG images, real endoscopic images of the realistic
head-model were used as the background. To prevent over-fitting to
the background images, more background images were generated by altering
the hue, saturation, and brightness of the original background images.
The intensity of the light coming from the virtual endoscope was also
varied to increase the amount of data. Robotic instruments are metallic;
therefore, they have a considerable specular reflection. To correctly
render background reflections in the shafts, the background images
were replicated in a cubic region around the shafts. The shafts themselves
are artificially rendered from a CAD model using physics-based shaders.
The surface roughness of the shafts was rendered using a normal map.

The shafts were rendered in random poses that are possible in endonasal
surgery. In detail, the poses were randomly chosen from the ranges
displayed in Table.~\ref{tab:dataset_range}. The rotation about
the shaft was excluded because of the rotational symmetry. We created
100000 images in total. Some examples of artificially generated images
are shown in Fig.~\ref{fig:sample_artificial_images}. The rendering
software automatically generates realistic CG images together with
appropriate bounding box information.

\begin{table}[tbh]
\centering

\caption{\label{tab:dataset_range} Range for the generation of the automatically
annotated data.}

\noindent\resizebox{\columnwidth}{!}{%

\textcolor{black}{}%
\begin{tabular}{cccccc}
\noalign{\vskip\doublerulesep}
Dimension & x {[}mm{]} & y {[}mm{]} & z {[}mm{]} & pitch {[}degree{]} & yaw {[}degree{]}\tabularnewline[\doublerulesep]
\hline 
\noalign{\vskip\doublerulesep}
\noalign{\vskip\doublerulesep}
Range & -20\textasciitilde 20 & -20\textasciitilde 20 & 10\textasciitilde 40 & 50\textasciitilde 90 & 0\textasciitilde 358\tabularnewline[\doublerulesep]
\noalign{\vskip\doublerulesep}
\end{tabular}

}
\end{table}

\subsection{Additional Training Methodologies}

The deep neural networks proposed in the next section have a limited
capability of generalizing from purely artificial images to real images.
However, it is difficult to obtain the precise pose of the instruments
in a real setup.

To balance this, we added to our dataset some real endoscopic images
and the manual annotation of their bounding boxes. For now, no pose
information was added to the real images. There is an ongoing research
effort in our group to create such a real dataset.

\subsection{Network architecture}

\begin{figure*}[tbh]
\centering

\def\svgwidth{2.0\columnwidth}

\scriptsize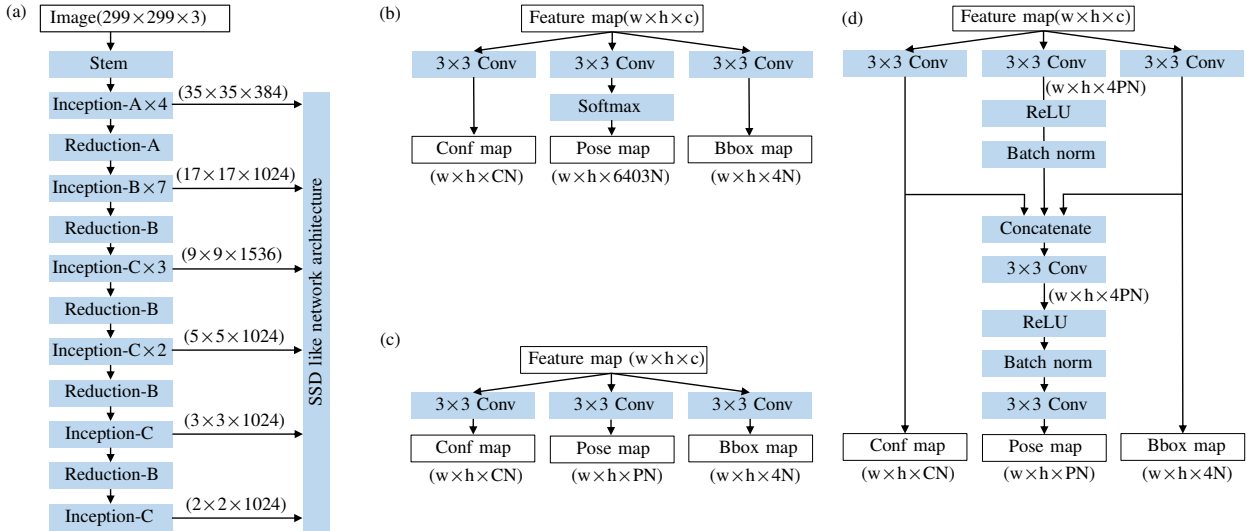

\caption{\label{fig:networl_model} The proposed network architecture. (a)
is the backbone of the SSD-like network created by extending Inception-V4
network. (b) is the SSD-like architecture proposed in SSD-6D, and
the pose output is in one-hot format. (c) and (d) are proposed SSD-like
architectures which output class confidence maps, bounding box maps,
and pose maps. (c) is the naive adjustment of the SSD-6D \cite{kehl2017ssd}
network to perform regression. (d) is an improved architecture to
estimate the pose more accurately. $P=5$ is the pose dimension.}
\end{figure*}

\subsubsection{Backbone network}

Our network can be regarded as an extension of the SSD-6D network
\cite{kehl2017ssd}. We use a network created by extending the InceptionV4
\cite{szegedy2017inception} as the backbone of the SSD \cite{liu2016ssd}.
The padding of 'Inception-B' is changed to obtain odd size feature
maps.

The input of the network is a $299\times299\times3$ $\left(\text{width}\times\text{height}\times\text{color}\right)$
image, resized from the high-definition images obtained from the endoscope.
Different from SSD-6D, our backbone network reduces the image to $2\times2$
feature maps instead of using $71\times71$ feature maps. This was
changed so that our network could appropriately detect the shafts,
given that they occupy a large portion of the image. The backbone
network is detailed in Fig.~\ref{fig:networl_model}, and its outputs
are six feature maps that are fed to the SSD-like network.

\subsubsection{Detection and pose estimation network}

In this work, we explore two architectures for instrument detection
and pose estimation. The first architecture is the naive adjustment
of the SSD-6D \cite{kehl2017ssd} network to perform regression, as
shown in Fig.~\ref{fig:networl_model} (b). The other architecture
is a more intricate network that feeds the class confidence prediction
output and bounding box prediction output to the pose estimation path
to increase the quality of the pose estimation, as shown in Fig.~\ref{fig:networl_model}
(c). We added ReLU activations and batch normalization to improve
the backpropagation. The concatenation of the bounding box prediction
path to the pose estimation path aims to give information on whether
the instruments are in the region. This facilitates the network learning
in that it outputs a valid pose value if the instruments are in the
region or outputs zero if the instruments are not in the region.

The networks output pose estimation maps, bounding box maps, and class
confidence maps. The bounding box maps and class confidence maps are
unchanged from SSD. The pose estimation outputs are normalized to
{[}-1, 1{]}, and the outputs are set to 0 if the instruments are not
on that location.

The shape of the output of the network is $(w_{i},h_{i},(C+4+P)N_{i})$
per feature map $i=1,2,3,4,5,6$ of the backbone network. The size
of each output is the same as the matching input feature map $\left(w_{i},h_{i}\right)$,
and $N_{i}$ is the number of bounding boxes per location that have
different aspect ratios. There is a pose estimation for each bounding
box $(w_{i},h_{i},P)$, an estimation of the parameters of the bounding
box itself $(w_{i},h_{i},4)$, and an estimation of the class confidence
$(w_{i},h_{i},C)$. Lastly, $P=5$ is the number of pose dimensions
and $C=2$ is the number of classes (instruments or background).

The detection and pose estimation network outputs many candidates
of bounding boxes and their corresponding pose estimation. The top
K high probability bounding-box candidates are chosen. Subsequently,
the final output of the bounding boxes is chosen using non-maximum
suppression. The final pose estimation is the pose corresponding to
the bounding boxes output by the non-maximum supression.

\subsection{Data augmentation}

At the training stage, several data augmentation methods are used.
For example, brightness, saturation, hue, and contrast of images are
randomly changed. Random additive noise is also added to each pixel
of each channel on some input images to increase the network's robustness
to different textures.

\subsection{Loss function}

Our loss functions were defined as

\[
L=\frac{1}{n}\left(L_{conf}+\alpha L_{bbox}+L_{pose}\right)
\]
where $n$ is the number of matched anchors. If the number of matched
anchors is 0, we set $n$ as 1. $L_{conf}$ is a softmax cross-entropy
loss for the class confidence map. We used the hard negative mining
method in $L_{conf}$ whose ratio between the negatives and the positives
is 3:1, similar to SSD \cite{liu2016ssd}. $L_{bbox}$ is a smooth
L1 loss for the bounding box estimation map. $L_{pose}$ is defined
as

\[
L_{pose}=\Sigma_{p}\beta_{p}L_{1}\left(\gamma(X_{p}^{pred}-X_{p}^{target})\right)
\]
where $p$ denotes the pose dimension and $L_{1}$ is a smooth L1
loss function. $X_{p}^{pred}$and $X_{p}^{target}$ are the pose estimation
and the correct pose of dimension $p$. $\alpha$, $\beta_{i}$, and
$\gamma$ are weighting parameters.

The Adam optimizer and a polynomial decay learning rate with power
2.0 were used for the smooth convergence of the loss.

\subsubsection{Loss switching for the training with real images}

Whenever the network was trained using real images, we assigned 0
to $L_{pose}$. This was required because the real images did not
have pose annotations.

\section{Evaluation}

Two experiments were conducted to evaluate our method. In the first
experiment, the quality of the pose estimation was evaluated on artificial
images. In the second experiment, we evaluated the proposed network
on real endoscopic images.

The same parameters were used for all architectures. Ninety thousand
artificial images were used as training data, and another 10000 images
were used as test data. The parameters were $\alpha=1.5$, $\gamma=5.0$,
$\beta_{i}=(1.0,1.0,2.0,2.0,2.0)$ in the order of (x, y, z, pitch,
yaw) making the weights larger in the dimensions that are difficult
to estimate from monocular images. Moreover, we used a batch size
of 32 during the training stage, and the detection threshold of the
SSD-like architecture was set to 0.5 IoU both in training and test
stages. The anchor ratio was the same as in SSD. The top 250 probability
detected candidates were chosen and they were input in the non-maximum
suppression. All the architectures were implemented in Python 3.6
using TensorFlow 1.13 and cuDNN 7.4 and executed in Ubuntu 18.04 with
a Quadro GV 100 graphics card.

\subsection{Evaluation on artificial images}

We first evaluated the error of the pose estimation using 10000 artificial
test images.We compared the performance of the naive adjustment of
SSD-6D, which we call Architecture-C, with the proposed architecture
that we call Architecture-D. Architecture-C and Architecture-D denote
the networks using (c) and (d) of Fig.~\ref{fig:networl_model},
respectively.

The evaluation metrics were the mean average precision (mAP), detection
rate, and pose error. The mAP was used as defined in PASCAL VOC 2010
\cite{pascal-voc-2010}. For the detection rate calculation, the instrument
was considered detected even if that single instrument was detected
as two instruments. Lastly, we evaluated the average pose error.

The results of the evaluation of the two architectures are shown in
Table.~\ref{tab:3d_pose_error}. Our results showed a considerable
increase in the quality of the estimation between Architecture-C (prior
method) and Architecture-D (proposed method). There were decreases
of 50\% , 45\% , and 63\% in the translation error, the pitch estimation
error, and the yaw estimation error, respectively.

To further improve the results of Architecture-D, we added uniform
random noise {[}-10.0, 10.0{]} to each pixel of each channel on a
training image with a 50\% probability. The result was '\emph{D with
noise}' Table.~\ref{tab:3d_pose_error}, which improved the pose
estimation.

To allow the network to generalize well to real images, we added 100
real images with manual bounding box annotation. We fed a mixture
of real images and artificial images to the network with a ratio of
2:9. The random uniform noise was also added. The result was '\emph{D
with noise and real}' in Table.~\ref{tab:3d_pose_error}. The detected
rate and pose estimation accuracy did not decrease in artificial images
even though the real images did not have any pose information.

To conclude this evaluation, Architecture-D ran on 30.4 fps which
is suitable for our real-time robot control. Comparing the best training
strategy of Architecture-D with the naive adjustment of SSD-6D (Architecture-C),
there was an error decrease of 55\% in position estimation, 64\% in
pitch, and 69\% in yaw.

\begin{table}[tbh]
\centering

\caption{\label{tab:3d_pose_error} Evaluation results on artificial test images.}

\noindent\resizebox{\columnwidth}{!}{%

\textcolor{black}{}%
\begin{tabular}{cccccccc}
\addlinespace
 & \multirow{2}{*}{mAP} & \multirow{2}{*}{Detected rate} & \multicolumn{5}{c}{Pose error}\tabularnewline\addlinespace
\addlinespace
\addlinespace
 &  &  & x {[}mm{]} & y {[}mm{]} & z {[}mm{]} & pitch {[}degree{]} & yaw {[}degree{]}\tabularnewline\addlinespace
\midrule
\addlinespace
\addlinespace
Architecture-C (prior method) & 0.958 & 0.937 & 1.51 & 1.41 & 1.64 & 3.97 & 13.58\tabularnewline\addlinespace
\addlinespace
\addlinespace
Architecture-D (proposed method) & 0.949 & 0.956 & 0.81 & 0.76 & 0.81 & 1.57 & 5.30\tabularnewline\addlinespace
\addlinespace
\addlinespace
D with noise & \textbf{0.965} & 0.985 & \textbf{0.65} & \textbf{0.63} & \textbf{0.75} & \textbf{1.44} & \textbf{4.26}\tabularnewline\addlinespace
\addlinespace
\addlinespace
D with noise and real & \textbf{0.965} & \textbf{0.988} & 0.66 & 0.64 & 0.77 & \textbf{1.44} & \textcolor{black}{4.33}\tabularnewline\addlinespace
\addlinespace
\end{tabular}

}

'D with noise' means the result of Architecture-D using random noise
data augmentation and 'D with noise and real' means the result of
Architecture-D using the random noise and real data.
\end{table}

\subsection{Preliminary evaluation on real images}

\begin{figure*}[t]
\centering

\def\svgwidth{2.0\columnwidth}

\scriptsize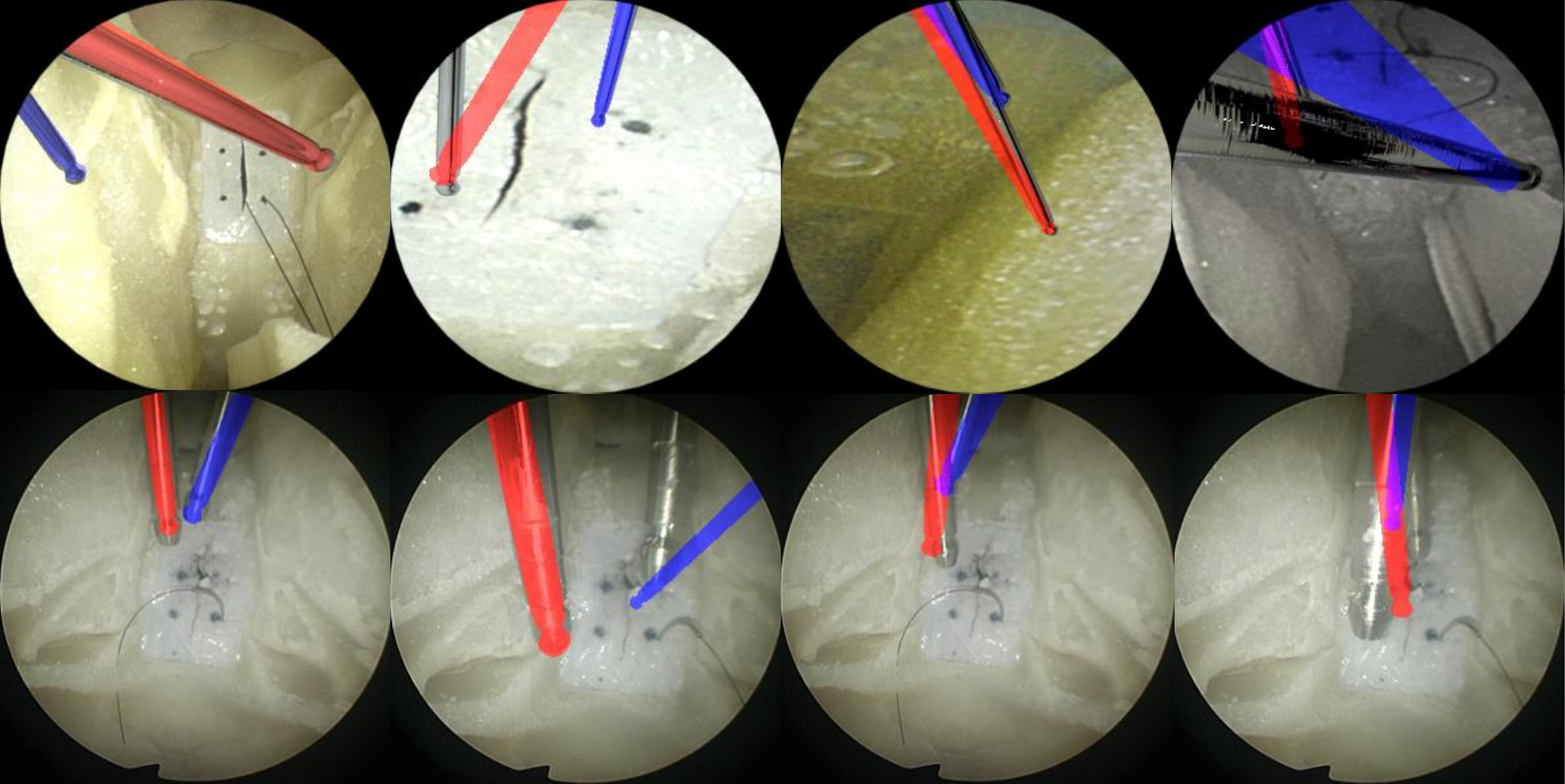

\caption{\label{fig:bounding_box_result} Visualization of the estimated pose.
The upper images are artificial images and the lower ones are real
endoscopic images.}
\end{figure*}

In the former experiment, the accuracy of the detection and pose estimation
were validated on artificial images and their automatic annotation.
In this section, an experiment was performed to evaluate the performance
of the '\emph{D with noise}' network and the '\emph{D with noise and
real}' network on real endoscopic images. However, given that we did
not have a dataset with the pose of real instruments, we evaluated
the detected rate and intersection over union (IoU).

To do so, first, we estimated the pose of the instruments in the real
images using the proposed network; then, we rendered artificial tools
on Blender using this pose information to generate the semantic segmentation
of the tools. We could then compare the estimated semantic segmentation
with manually annotated images to obtain the IoU. In total, 40 real
endoscopic images were evaluated; 20 of them had no occlusion and
the others had heavily occluded instruments.

The result of this evaluation is shown in Table.~\ref{tab:application_performance}
and Fig.~\ref{fig:bounding_box_result}. It is important to note
that the absolute value of the IoU, in this case, is not always a
good indicator of pose error, as shown in Table.~\ref{tab:3d_pose_error}.
Nonetheless, the relative value of the IoU evaluated on real images
with respect to the IoU evaluated on artificial images shows that
the method can also be applied to real images to some extent. This
result is interesting because there was no pose annotation for the
real images used during the training of the network and the network
still learned how to estimate the instrument's pose. There is an ongoing
effort in our group to improve the quality of our dataset to include
real images with full annotation.

\begin{table}[tbh]
\centering

\caption{\label{tab:application_performance} Performance comparison over artificial
images and real endoscopic images.}

\noindent\resizebox{\columnwidth}{!}{%

\textcolor{black}{}%
\begin{tabular}{cccccc}
\noalign{\vskip\doublerulesep}
 &  & \multicolumn{2}{c}{non occluded images} & \multicolumn{2}{c}{partly occluded images}\tabularnewline[\doublerulesep]
\hline 
\noalign{\vskip\doublerulesep}
\noalign{\vskip\doublerulesep}
 &  & Detected rate & IoU & Detected rate & IoU\tabularnewline[\doublerulesep]
\noalign{\vskip\doublerulesep}
\noalign{\vskip\doublerulesep}
Artificial & D with noise and real & 0.950 & 0.494 & 0.850 & 0.369\tabularnewline[\doublerulesep]
\noalign{\vskip\doublerulesep}
\noalign{\vskip\doublerulesep}
\multirow{2}{*}{Real} & D with noise & 0.700 & 0.197 & 0.700 & 0.195\tabularnewline[\doublerulesep]
\noalign{\vskip\doublerulesep}
\noalign{\vskip\doublerulesep}
 & D with noise and real & 0.950 & 0.471 & 0.825 & 0.299\tabularnewline[\doublerulesep]
\noalign{\vskip\doublerulesep}
\end{tabular}

}
\end{table}

\section{Conclusions}

In this paper, we proposed a deep-learning methodology to estimate
the pose of the shaft of surgical instruments in monocular endoscopic
images. To do so, we extended the SSD-6D \cite{kehl2017ssd} network.
We also created a dataset composed of artificially rendered images
with the automatic annotation of the instruments' bounding boxes and
poses.

In the context of our study, we could considerably decrease pose estimation
error (55\% in position estimation, 64\% in pitch, and 69\% in yaw)
with our proposed architecture in an experiment using artificial images.
Moreover, in a preliminary experiment using real images, we showed
that the network could generalize from real images to some extent.
Further work must be done to improve our training dataset and add
real images with pose annotation.

This is an important first step in our goal to the online calibration
of the instrument's shafts, aiming at reliable collision avoidance
for robot-aided surgical procedures in constrained workspaces. Our
method should be applicable for other instrument designs as long as
the instrument has a prominent shaft. In future works, we intend to
utilize sequential video information to increase the robustness of
our predictions.

\balance

\appendices{}

\bibliographystyle{IEEEtran}
\bibliography{bib/ral}

\end{document}